\documentclass[runningheads]{llncs}
\usepackage{amsmath}
\usepackage[utf8]{inputenc} 

\usepackage[T1]{fontenc}    
\usepackage{hyperref}       
\usepackage{url}            
\usepackage{booktabs}       
\usepackage{amsfonts}       
\usepackage{nicefrac}       
\usepackage[expansion=false]{microtype}
\usepackage{graphicx}
\usepackage{longtable}
\graphicspath{ {./images/} }
\usepackage{fancyvrb}
\usepackage{tikz}
\usepackage{subcaption}
\usepackage{pgfplots}
\usepackage{tabularx}

\usepgfplotslibrary{groupplots}
\pgfplotsset{compat=1.17}
\usetikzlibrary{shapes.geometric, arrows, arrows.meta}

\begin{document}

\title{Reinforcement Learning Environment with LLM-Controlled Adversary in
D\&D 5th Edition Combat}

%
\author{Joseph Emmanuel DL Dayo\inst{1}\orcidID{0009-0007-6943-2003} \and
Michel Onasis S. Ogbinar\inst{2}\orcidID{0009-0008-0231-4173}
\and
Prospero C. Naval Jr.\orcidID{0000-0001-7140-1707}\inst{3}}
%

%
\institute{University of the Philippines, Diliman, Quezon City, Philippines  \\
\email{jddayo@up.edu.ph}\\
\and
\email{msogbinar@up.edu.ph}\\
\and
\email{pcnaval@up.edu.ph}}

\authorrunning{Dayo, Ogbinar and Naval}
\titlerunning{RL Environment with LLM-Controlled Adversary in
D\&D 5E Combat}
\maketitle

\begin{abstract}
The objective of this study is to design and implement a reinforcement learning (RL) environment using D\&D 5E combat scenarios to challenge smaller RL agents through interaction with a robust adversarial agent controlled by advanced Large Language Models (LLMs) like GPT-4o and LLaMA 3 8B. This research employs Deep Q-Networks (DQN) for the smaller agents, creating a testbed for strategic AI development that also serves as an educational tool by simulating dynamic and unpredictable combat scenarios. We successfully integrated sophisticated language models into the RL framework, enhancing strategic decision-making processes. Our results indicate that while RL agents generally outperform LLM-controlled adversaries in standard metrics, the strategic depth provided by LLMs significantly enhances the overall AI capabilities in this complex, rule-based setting. The novelty of our approach and its implications for mastering intricate environments and developing adaptive strategies are discussed, alongside potential innovations in AI-driven interactive simulations. This paper aims to demonstrate how integrating LLMs can create more robust and adaptable AI systems, providing valuable insights for further research and educational applications.
\end{abstract}

\section{Introduction}
The field of Artificial Intelligence (AI) has witnessed transformative advancements with the advent of Large Language Models (LLMs). These models, trained on extensive text corpora, have demonstrated remarkable capabilities in natural language processing tasks such as text generation, question answering, and machine translation. However, the potential of LLMs extends far beyond these traditional language-centric applications.

Our research explores a novel frontier: leveraging the strengths of LLMs in complex, strategic decision-making environments. Specifically, we propose integrating advanced language models into a Reinforcement Learning (RL) framework, using the intricate and rule-intensive domain of Dungeons \& Dragons 5th Edition (D\&D 5E) combat scenarios. This unique combination aims to push the boundaries of AI's capabilities in mastering complex, rule-based systems.

Strategic games have long served as benchmarks for AI development, offering controlled yet challenging environments that test an agent's ability to make complex decisions under uncertainty. Recent breakthroughs, such as DeepMind's AlphaGo in Go and OpenAI's agents in Dota 2, underscore the potential of AI in this domain. However, these successes often rely on specialized architectures tailored to specific game mechanics. We posit that LLMs, with their deep understanding of human language and demonstrated ability to follow intricate instructions, could excel in highly structured, language-heavy strategic environments. D\&D 5E, with its extensive rulebooks and dynamic gameplay, presents an ideal testbed. Its combat scenarios require interpreting complex rules, understanding context, and formulating adaptive strategies—challenges that align closely with the strengths of LLMs.

Our primary goal is to design and implement an RL environment that harnesses LLMs for strategic decision-making in D\&D 5E combat scenarios. Key objectives include creating an adversarial agent controlled by an LLM (e.g., GPT-4o or LLaMA 3 8B) that can interpret game states, understand rules, and make strategic decisions; training smaller RL agents to compete against this LLM-driven adversary, fostering a competitive environment that drives the development of sophisticated strategies; evaluating the effectiveness of LLMs in enhancing strategic decision-making processes within the RL framework; and exploring the potential of this setup as a robust testbed for strategic AI development.

The novelty and significance of our work lie in several areas. We incorporate state-of-the-art language models like GPT-4o and LLaMA 3 into an RL framework, specifically for strategic decision-making. By using D\&D 5E, we challenge AI to master an environment with intricate, language-heavy rules—a scenario more reflective of real-world complexities than many current game-based testbeds. Our approach casts the LLM as a dynamic, adaptive opponent, providing a level of unpredictability and strategic depth often lacking in traditional AI adversaries. This research offers valuable insights into AI's ability to understand and operate within complex, rule-based systems, with implications far beyond gaming. Furthermore, our work opens avenues for innovative AI applications, particularly in areas requiring strategic decision-making, such as educational simulations, policy planning, and complex system management.

RL has been instrumental in developing AI agents capable of learning and improving through interaction with their environments. Notable successes include DeepMind's AlphaGo\cite{silver2016mastering}, which mastered the game of Go through self-play and RL, and OpenAI's Dota 2 agents\cite{berner2019dota}, which demonstrated the ability to collaborate and compete at a high level of play.

Existing literature on the application of AI and ML in games highlights significant strides in using RL and LLMs for game simulations. For instance, LLM research that investigate the capabilities of LLMs like GPT-4o in playing text games\cite{tsai2023can}, underscoring the strengths and limitations of these models in understanding and interacting with complex environments. This study is directly relevant to our research on employing LLMs in D\&D 5E combat scenarios. Another introduces fundamental RL concepts in two-player zero-sum games\cite{phillips2021reinforcement} and presents novel agents using joint action DQN, which offer valuable insights for developing smaller RL agents in our framework. Next is explored the integration of language game paradigms with RL\cite{van2020re}, emphasizing potential breakthroughs in emergent communication. Their work aligns with our goal of using sophisticated language models to enhance strategic decision-making in RL environments. Studies also demonstrate how LLMs can guide the pretraining of RL agents\cite{du2023guiding}, enhancing their strategic capabilities. The integration of LLMs in text-based games is explored in "Language Understanding for Text-based Games using Deep Reinforcement Learning," which illustrates the joint learning of state representations and action policies using game rewards. Lastly, innovative approaches for RL agents in complex, strategic game settings like D\&D 5E, addressing the challenges of infinite action spaces and collaborative play\cite{martin2018dungeons}.

Despite the impressive advancements in AI and ML applications in game environments, several gaps remain in the current research, particularly regarding the use of LLMs in complex, rule-based role-playing games (RPGs) like D\&D 5E. Existing studies have predominantly focused on simpler game environments or those with well-defined action spaces, such as board games, real-time strategy games, and text-based simulations. There is limited research on the integration of LLMs with RL frameworks to create adaptive and strategic AI agents in these more intricate settings.

Most existing research has treated LLMs and RL separately, without exploring their combined potential. This has left a noticeable gap in understanding how LLMs can be applied to control adversarial agents within RPGs, where the action space is vast and the game dynamics are more complex. The capabilities of LLMs to simulate dynamic and unpredictable combat scenarios, and their potential to enhance strategic decision-making in these environments, remain under-explored.

\section{Integrating LLMs into Reinforcement Learning}

The theoretical framework underpinning this research integrates principles from RL and LLMs, leveraging advancements in each field to address the challenges of strategic decision-making in complex, rule-based environments such as D\&D 5E combat scenarios.

At its core, RL involves training agents to make sequential decisions by maximizing cumulative rewards through interactions with their environment\cite{sutton1999reinforcement}. This method has shown remarkable success in various game-playing AI applications. Key concepts in RL include Q-learning, policy gradients, and value functions, which guide the agent's decision-making process. Recent advancements in deep RL, such as DQNs have significantly enhanced the ability of AI agents to learn from complex environments\cite{mnih2013playing}. DQNs, in particular, have demonstrated the capability to master intricate game dynamics by learning optimal strategies through trial-and-error interactions.

On the other hand, LLMs provide sophisticated natural language understanding and generation capabilities. Based on transformer architectures, these models are trained on extensive datasets to comprehend and produce human-like text\cite{openai2024gpt4}. Their development has pushed the boundaries of what LLMs can achieve, showcasing their potential in diverse applications beyond traditional NLP tasks. In the context of game environments, LLMs facilitate more nuanced interactions and strategic planning by enabling AI systems to interpret complex game states, understand rules, and generate contextually appropriate actions.

The integration of RL and LLMs offers a novel approach to developing AI systems that can adeptly manage the complexities of rule-based games, particularly RPGs. This synergy leverages RL's adaptive learning capabilities with LLMs' language understanding proficiency. It can be seen that LLMs can be used to understand character actions and narrative contexts in case of RPGs\cite{louis2018deep}.

Furthermore, LLMs can enhance the strategic depth of RL agents by incorporating human-like reasoning and adaptability into their decision-making processes. In D\&D 5E combat scenarios, for example, an LLM can interpret the current battle state, understand the capabilities and weaknesses of each character, and generate a range of strategic actions\cite{zhu2023fireball}. The RL agent then evaluates these actions based on their potential long-term rewards, choosing the most promising strategy. This combination allows the AI to make decisions that are not only optimal in terms of game mechanics but also narratively coherent and tactically sophisticated.

A key innovation in this research is the implementation of continuous learning mechanisms. Here, RL agents adapt their strategies based on interactions with LLM-controlled adversaries, fostering an environment of ongoing improvement and strategic evolution. This approach allows the AI system to learn from each encounter\cite{carta2023grounding}, refining its understanding of game dynamics and adapting its tactics in response to diverse and unpredictable opponent behaviors.

\section{Methodology}

This research utilizes a sophisticated D\&D 5E game engine designed specifically for AI-related research\cite{Dayo2024}. The core functionality of this codebase includes managing game sessions, handling combat mechanics, and integrating AI agents to interact within the game environment. This setup provides a robust platform for training and testing AI agents within a complex and dynamic environment.

\begin{figure*}[h]
\centering
\tikzstyle{block} = [rectangle, draw, fill=blue!20, 
    text width=4em, text centered, rounded corners, minimum height=3em, font=\tiny] 
\tikzstyle{line} = [draw, -Latex]

\begin{tikzpicture}[auto,scale=0.8, transform shape]
    \node [block] (players) at (0,0) {Players \& NPCs};
    \node [block] (initiative) at (2,0) {Initiative Order};
    \node [block] (actions) at (4,0) {Actions: Attack, Spell, etc.};
    \node [block] (rules) at (7,0) {Apply Rules: AC, Saving Throws};
    \node [block] (health) at (10,0) {Update Health, Effects};
    \node [block] (nextturn) at (12,0) {Next Turn};
    
    \draw[line] (players) -- (initiative);
    \draw[line] (initiative) -- (actions);
    \draw[line] (actions) -- (rules);
    \draw[line] (rules) -- (health);
    \draw[line] (health) -- (nextturn);
    \draw[line] (nextturn.south) to[bend left=15] (initiative.south);
    \draw[line] (health.north) to[bend right=15] (actions.north);
\end{tikzpicture}
\caption{D\&D 5E Combat Flow}
\label{fig:Picture5.png}
\end{figure*}
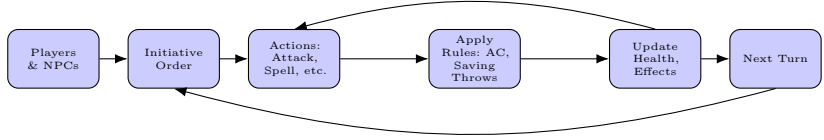

The simulated environment is based on D\&D 5E combat scenarios, As shown in Figure \ref{fig:Picture5.png}, incorporating a detailed state space, action space, and roles for different AI agents. The environment, which implements a subset of the D\&D 5E rules based on the SRD (System Reference Document)\cite{wizards2024dnd}, is managed by a comprehensive game engine that defines various entities, such as players and NPCs, with specific attributes and behaviors.

\subsubsection{State and Action spaces}

The state space includes detailed representations of the battlefield, character positions, health statuses, environmental features, and available actions.  The action space encompasses all possible actions that an agent can take, including movement, combat actions, and special abilities. In the rules of D\&D 5E there is a resource constrained set of actions players can take during their turn as well as defined "reactions" that can be taken when certain conditions are met. However the variety of these actions can be very large, (i.e. while players can only generally take one attack action, there is a large variations of means to execute this, for example a choice needs to be made on which weapon to use and which target to hit). This becomes even more complicated in the case of spell-casting where a choice needs to be made on which spell to use out of hundreds of possible types of spells. We have to note that range of valid actions can vary widely depending on the state of the environment. To assist with this, the environment already pre-generates the allowable valid actions at that state for the current player and returns it as part of the observation meta-state. The DQN agent would therefore only need to evaluate each valid state-action pair and pick the one with the highest value. For this study we limit the types of spells and weapons available and are fixed for each class.

\tikzstyle{block} = [rectangle, draw, fill=blue!20, 
    text width=6em, text centered, rounded corners, minimum height=2em]
\tikzstyle{line} = [draw, -latex']

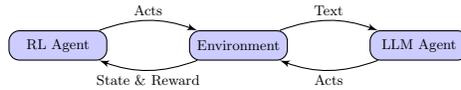
\begin{figure*}[h]
\centering
\begin{tikzpicture}[node distance = 7cm, auto, scale=0.6, transform shape]
    \node [block] (rlagent) {RL Agent};
    \node [block, right of=rlagent, node distance=4cm] (environment) {Environment};
    \node [block, right of=environment, node distance=4cm] (llmagent) {LLM Agent};
    \path [line] (rlagent) edge[bend left=20] node {Acts} (environment);
    \path [line] (environment) edge[bend left=20] node [below] {State \& Reward} (rlagent);
    \path [line] (environment) edge[bend left=20]node {Text} (llmagent);
    \path [line] (llmagent) edge[bend left=20] node [below] {Acts} (environment);
\end{tikzpicture}
\caption{RL \& LLM Agent and Environment Interaction}
\label{fig:Picture1.png}
\end{figure*}

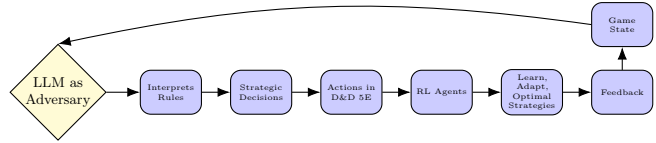
\begin{figure*}[h]
\centering
\tikzstyle{block} = [rectangle, draw, fill=blue!20, 
    text width=3.5em, text centered, rounded corners, minimum height=3em, font=\tiny]
\tikzstyle{line} = [draw, -Latex]
\tikzstyle{decision} = [diamond, draw, fill=yellow!20, 
    text width=4.5em, text badly centered, node distance=2cm, inner sep=0pt]

\begin{tikzpicture}[auto, scale=0.6, transform shape]
    \node [decision] (start) at (0,0) {LLM as Adversary};
    \node [block] (interpret) at (2.5,0) {Interprets Rules};
    \node [block] (strategic) at (4.5,0) {Strategic Decisions};
    \node [block] (actions) at (6.5,0) {Actions in D\&D 5E};
    \node [block] (rlagents) at (8.5,0) {RL Agents};
    \node [block] (learning) at (10.5,0) {Learn, Adapt, Optimal Strategies};
    \node [block] (feedback) at (12.5,0) {Feedback};
    \node [block] (gamestate) at (12.5,1.5) {Game State};
    
    \draw[line] (start) -- (interpret);
    \draw[line] (interpret) -- (strategic);
    \draw[line] (strategic) -- (actions);
    \draw[line] (actions) -- (rlagents);
    \draw[line] (rlagents) -- (learning);
    \draw[line] (learning) -- (feedback);
    \draw[line] (feedback) -- (gamestate);
    \draw[line] (gamestate.west) to[bend right=10] (start.north);
\end{tikzpicture}
\caption{RL \& LLM Integration}
\label{fig:Picture4.png}
\end{figure*}

We implemented various scenarios to measure the performance with respect to increasing complexity of the environment. Initially we limit the types of players to only the fighter class with the elf race, after which we increase the complexity by allowing the environment to select randomly from different classes like the mage, cleric and rogue. As shown in Figure \ref{fig:Picture1.png}, the AI agents include an adversarial agent (A\_LLM), controlled by an LLM, which is capable of interpreting complex game states and making high-level strategic decisions, and reinforcement learning agents (A\_RL), which are smaller, specialized agents trained using RL algorithms like DQN, operating within a constrained action space, as shown in Figure \ref{fig:Picture4.png}.

\vspace{-10mm} 

\subsubsection{Large Language Models}

The study employs various AI models and algorithms to explore their effectiveness in strategic decision-making within D\&D 5E combat scenarios. The LLMs used in this study include GPT\-4o-mini, chosen for its combination  of low-cost, advanced natural language understanding and generation capabilities, making it ideal for interpreting complex rules and generating strategic actions. LLaMA 3.1 8B Instruct was selected for its balance of computational efficiency and language processing capabilities, and Mistral 7B Instruct 0.3 was included for comparative analysis, focusing on its strengths in language generation. What makes D\&D a good case for LLMs is that D\&D 5E has been around for decades and have numerous online text corpora including its rules and gameplay that would have been used in the training sets of these LLMs. 

We note that LLMs only accept input in the form of tokens from the domain of human language, hence a prompting strategy is used to convert the observed state and actions into the form of a prompt. An example is shown in Fig. \ref{fig:prompting}, which show the prompt template used. The state of the map which includes obstacles and positions of the players are converted to an ASCII readable map, a short description of player and enemy states, and sent along with the available actions that the LLM is instructed to choose from. Finally, the response text is then parsed to obtain the selected action.

\begin{figure}[h]
    \centering
\begin{multicols}{2}
\tiny
\begin{verbatim}
prompt: -------------------------------
We are playing a game of Dungeons and
Dragons
5th Edition. It is current your turn and
you play 
as a hero character denoted by P
(a level 2 rogue).
And you have an enemy donoted
by E (a level 2 wizard)
which you must defeat. 
Your health is at [83.33333333]%
specifically 15/18 
Your Enemies health is at [100.]%
Your current conditions are:
Your enemies current conditions are:
You have the following available actions
and movement available:

Available movement: [25]ft
Available actions: 1
Bonus actions: 1
Reactions: 1

Here is a rough sketch of the map that
considers line of sight to the enemy.
Here is the map:
____________
____________
____________
____________
____________
___.......__
___...P...__
___.E.*...__
___.~~ ~..__
___~~   ..__
___~~   ..__
____________
areas with no characters are represented
by a dot (.)
the hero character is represented by
a (P)
the enemy character is represented by
an (E)
Allies or Party Members are represented by an (A)
Neutral characters are represented by
a question mark (?)
areas outside of the map are represented by a hash (_),
you
cannot move to areas with _
areas with obstacles are represented by an asterisk (*)
areas with a barrel are represented by an (o).
These provide half-cover if right behind it and
attacks are comming from the other side.
areas with water are represented by a tilde (~) and
are difficult terrain
areas that the player can't see are just blanks/space
Each tile of the map is 5ft by 5ft.

Here are the available actions you can take,
please choose the number corresponding to the action:
0: end my turn
1: attack enemy with ranged weapon: dagger
2: attack enemy with ranged weapon: dagger
3: dash action
4: dash as bonus action
5: disengage action
6: disengage as bonus action action
7: dodge action
8: move 5ft up and to the left
9: move 5ft to the left
10: move 5ft down and to the left
11: move 5ft up
12: move 5ft up and to the right
13: move 5ft to the right
14: move 5ft down and to the right
15: go prone

Please choose the number corresponding to the action
you would like to take.
Provide your answer using the format, starting with
the desired number choice, followed by the
colon and the action.
1: attack enemy with ranged weapon
Just provide the action choice, no need to explain.

---------------------------------
{'action': 9}
response time: 1.4391560554504395
\end{verbatim}
\end{multicols}

    \caption{Sample LLM prompt for an environment state}
    \label{fig:prompting}
\end{figure}

The response text provided by the LLMs is not always stable (i.e. not following the instructions), and in some cases the response is difficult to use. In the case of GPT-4o we employ the "tools" functionality to force the response into JSON format. For the other LLMs we fallback to picking a random action when this happens.

\subsubsection{Training DQN with LLM}

We train our DQN with an environment where the adversarial agent is controlled by an LLM. The DQN updates its Q-values using the standard Bellman equation:

\[
Q(s, a) \leftarrow Q(s, a) + \alpha \left( r + \gamma \max_{a'} Q(s', a') - Q(s, a) \right),
\]

where \(s\) is the current state, \(a\) is the DQN agent's action, \(r\) is the reward received, \(s'\) is the next state, and \(a'\) is the next action. The LLM acts purely as an adversarial agent, interpreting the current state of the environment \(s\) and selecting adversarial actions based on its understanding of the game rules.

The DQN agent receives feedback in the form of rewards \(r\) based on its interactions with the environment, which is partially driven by the LLM's decisions. Over time, the DQN learns to optimize its actions against the LLM-controlled adversary to maximize cumulative rewards.

\subsubsection{DQN Network Architecture}

The model architecture (Fig \ref{fig:dqn-network}) processes the environments state which consists of a 2D map of the 7x7 viewport with 16 channels (Map tile features like terrain, object features) with other player states as embeddings (e.g. weapon, DnD Class) including the  action, with the final fully-connected linear layers in order to predict the value for the state action pair.

\tikzstyle{startstop} = [rectangle, rounded corners, minimum width=1cm, minimum height=0.8cm,text centered, draw=black, fill=red!30]
\tikzstyle{io} = [trapezium, trapezium left angle=70, trapezium right angle=110, minimum width=1cm, minimum height=0.8cm, text centered, draw=black, fill=blue!30]
\tikzstyle{process} = [rectangle, minimum width=1cm, minimum height=0.8cm, text centered, text width=3cm, draw=black, fill=orange!30]
\tikzstyle{arrow} = [thick,->,>=stealth]

\tikzstyle{startstop} = [rectangle, rounded corners, minimum width=1cm, minimum height=0.8cm,text centered, draw=black, fill=red!30]
\tikzstyle{io} = [trapezium, trapezium left angle=70, trapezium right angle=110, minimum width=1cm, minimum height=0.8cm, text centered, draw=black, fill=blue!30]
\tikzstyle{arrow} = [thick,->,>=stealth]

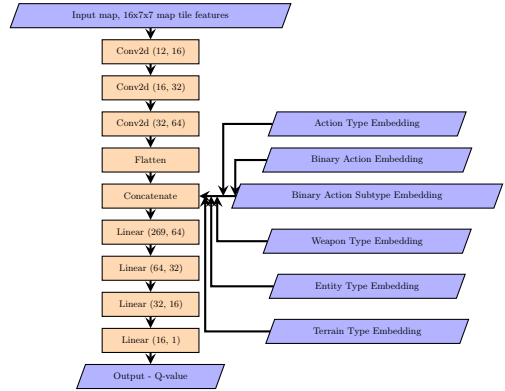
\begin{figure}[h]
\centering

\begin{tikzpicture}[node distance=1.2cm, scale=0.4, transform shape]

\node (start) [io] {Input map, 16x7x7 map tile features};
\node (conv1) [process, below of=start] {Conv2d (12, 16)};
\node (conv2) [process, below of=conv1] {Conv2d (16, 32)};
\node (conv3) [process, below of=conv2] {Conv2d (32, 64)};
\node (flatten) [process, below of=conv3] {Flatten};
\node (concat1) [process, below of=flatten] {Concatenate};
\node (fc1) [process, below of=concat1] {Linear (269, 64)};
\node (fc2) [process, below of=fc1] {Linear (64, 32)};
\node (fc3) [process, below of=fc2] {Linear (32, 16)};
\node (output) [process, below of=fc3] {Linear (16, 1)};

\node (emb1) [io, right of=conv3, xshift=6cm] {Action Type Embedding};
\node (emb2) [io, below of=emb1] {Binary Action Embedding};
\node (emb3) [io, below of=emb2] {Binary Action Subtype Embedding};
\node (emb4) [io, below of=emb3, yshift=-0.3cm] {Weapon Type Embedding};
\node (emb5) [io, below of=emb4, yshift=-0.3cm] {Entity Type Embedding};
\node (emb6) [io, below of=emb5, yshift=-0.3cm] {Terrain Type Embedding};

\node (foutput) [io, below of=output] {Output - Q-value};

\draw [arrow] (start) -- (conv1);
\draw [arrow] (conv1) -- (conv2);
\draw [arrow] (conv2) -- (conv3);
\draw [arrow] (conv3) -- (flatten);
\draw [arrow] (flatten) -- (concat1);
\draw [arrow] (concat1) -- (fc1);
\draw [arrow] (fc1) -- (fc2);
\draw [arrow] (fc2) -- (fc3);
\draw [arrow] (fc3) -- (output);
\draw [arrow] (output) -- (foutput);

\draw [arrow] (emb1) -| ([xshift=0.8cm]concat1.east);
\draw [arrow] (emb2) -| ([xshift=1.2cm]concat1.east);
\draw [arrow] (emb3) -- (concat1.east);
\draw [arrow] (emb4) -| ([xshift=0.6cm]concat1.east);
\draw [arrow] (emb5) -| ([xshift=0.4cm]concat1.east);
\draw [arrow] (emb6) -| ([xshift=0.2cm]concat1.east);

\end{tikzpicture}
    \caption{DQN Network}
    \label{fig:dqn-network}
\end{figure}

\subsubsection{Environment Setup}

\begin{figure}
    \begin{minipage}{0.20\textwidth}
        \centering
        \includegraphics[width=\textwidth]{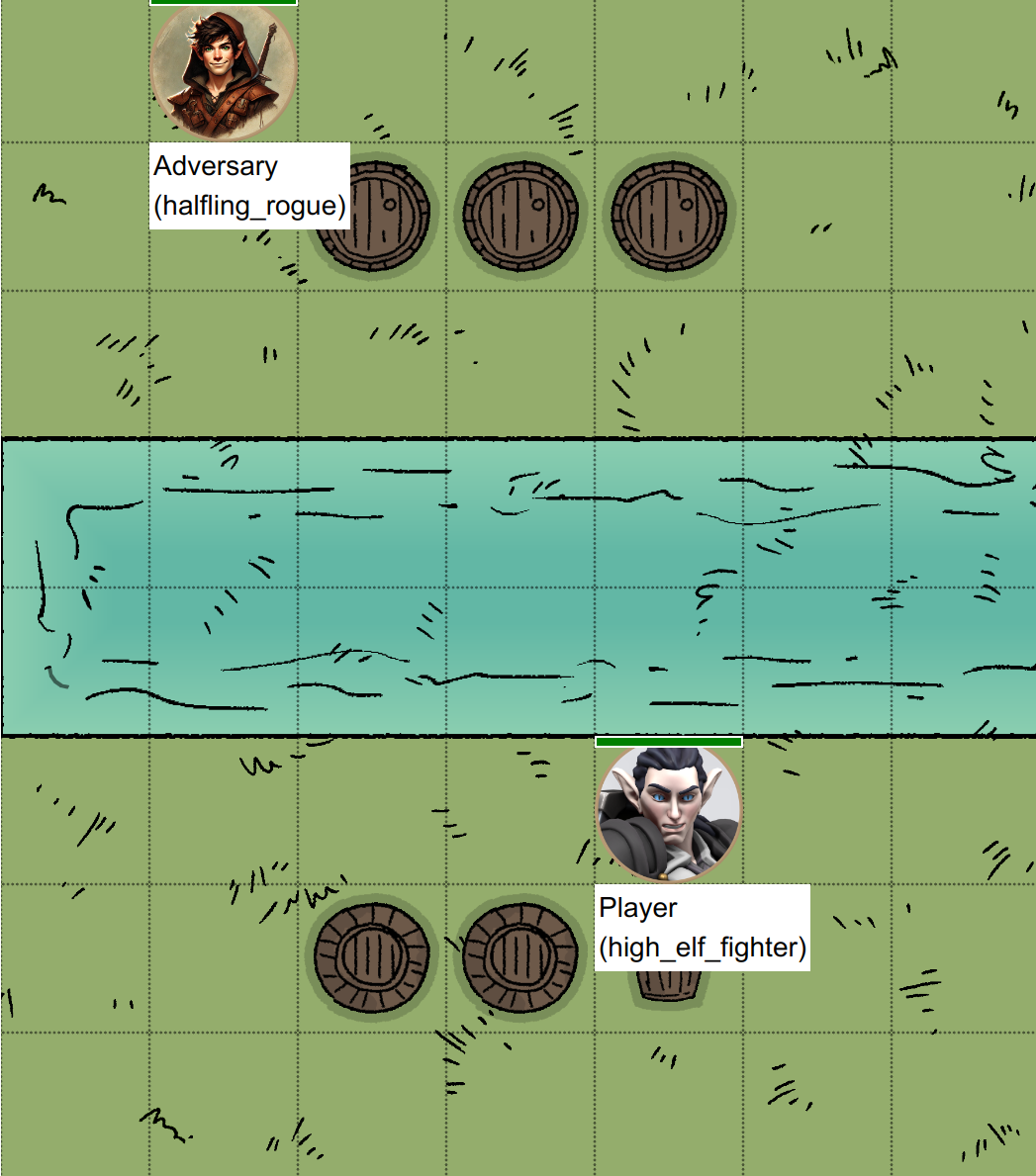} 
    \end{minipage}
    \begin{minipage}{0.15\textwidth}
        \begin{verbatim}
.E.....
..ooo..
.......
wwwwwww
wwwwwww
....P..
..ooo..
.......
        \end{verbatim}
    \end{minipage}
        \begin{minipage}{0.20\textwidth}
        \centering
        \includegraphics[width=\textwidth]{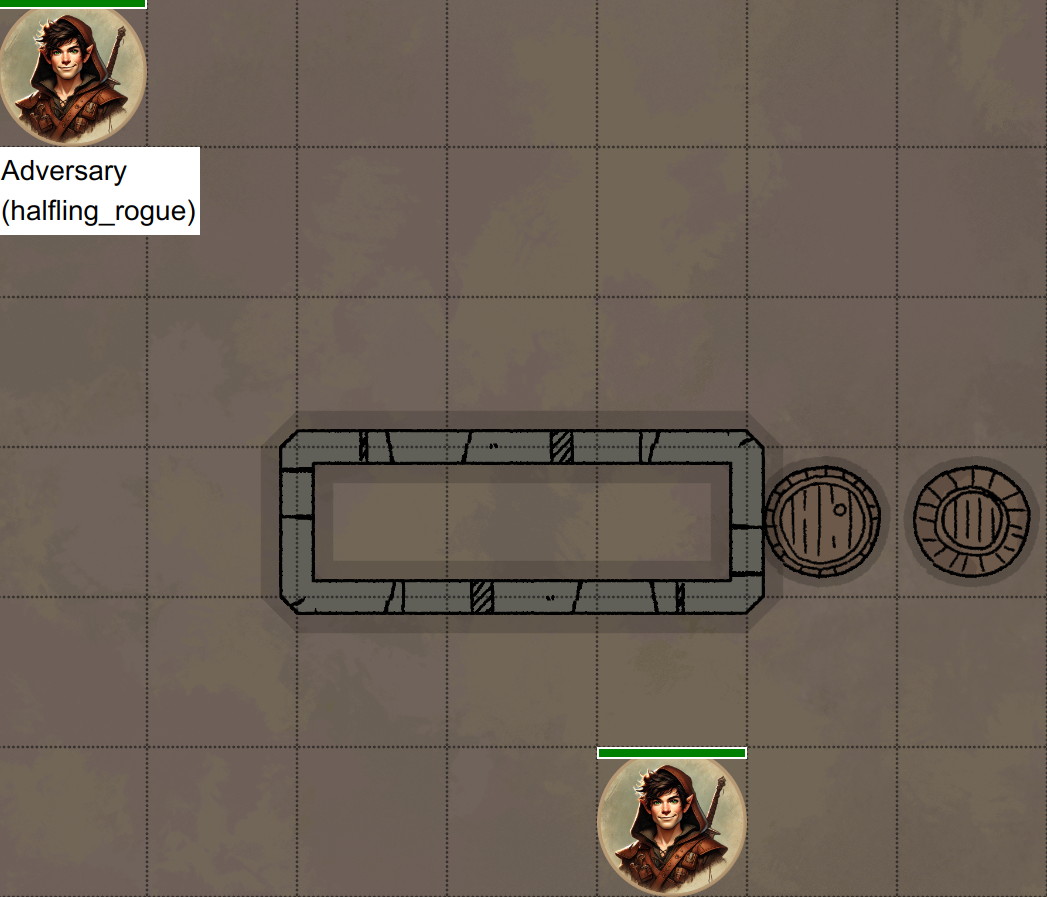} 
    \end{minipage}
    \begin{minipage}{0.15\textwidth}
        \begin{verbatim}
E......
.......
.......
..###oo
.......
....P..
        \end{verbatim}
    \end{minipage}

\tiny
\centering
\begin{tabular}{l l l l p{3.5cm} p{3.5cm}}
\toprule
\textbf{Name} & \textbf{Race} & \textbf{Class} & \textbf{HP} & \textbf{Abilities} & \textbf{Equipment} \\
\midrule
Shor & Mountain Dwarf & Cleric 2 & 16 & STR 14, DEX 10, CON 16, INT 10, WIS 16, CHA 12 & Dagger, Scale Mail, Shield, Warhammer \\
Belly & Lightfoot Halfling & Rogue 2 & 18 & STR 11, DEX 20, CON 16, INT 11, WIS 12, CHA 17 & 2x Dagger, Torch, Leather Armor \\
Gom & High Elf & Fighter 2 & 24 & STR 12, DEX 20, CON 16, INT 16, WIS 12, CHA 11 & Rapier, Longbow, Leather Armor, Shield \\
Crys & High Elf & Wizard 2 & 14 & STR 10, DEX 15, CON 14, INT 18, WIS 12, CHA 8 & Dagger \\
\bottomrule
\end{tabular}
\caption{Characters}
\label{fig:characters}
\end{figure}

The experimental methodology focuses solely on 1-on-1 combat scenarios. We start with a tournament setup where one Level 2 Elf Fighter controlled by an RL agent competes against an Elf Fighter controlled by an LLM. After that, we then setup a tournament where the character classes (level 2) are randomized for each round (Fig \ref{fig:characters}). We choose level 2 characters to strike a balance where a player doesn't get killed outright in one round, while also not being high enough where matches can take too long to complete.

The evaluation involves conducting 30 fights per match-up, with each fight continuing until one side dies or both sides survive (resulting in a tie). We measure the performance of each of the agents by obtaining the total victories accumulated. This methodology provides a comprehensive evaluation of the AI models' strategic decision-making and adaptive learning capabilities within the complex, rule-based environment of D\&D 5E combat scenarios.

As for the maps, we developed (4) different types of maps (Fig \ref{fig:characters}) that include obstacles like walls, difficult terrain as well as objects that can be used as cover. One of the maps is a control map where it is a plain map with no obstacles. These maps are then picked randomly at the start of each episode.

\subsubsection{Adversarial Agents}

Aside from LLM-based adversaries, there is also a simple custom rules-based AI that serves as a baseline for testing the DQN agent. This custom AI operates under a set of predefined mechanics: if an attack is possible, it executes the attack; if not, it moves closer to the hero and then attempts an attack if within range. Both players also have access to other abilities like, second wind and action surge\cite{wizards2024dnd} that depend on correct timing to be used effectively. To evaluate the effectiveness of the DQN agent, we conducted a series of tests comparing the custom AI against the DQN agent.

In the case of GPT-4o, in order to reduce the expense and limitations of its paid tier we employed an approach where we only use GPT-4o for major decisions (like those costing an action), while movement and minor decisions are off-loaded to a secondary LLM like LLama 3.

\subsubsection{Reward Function}

For the rewards, a value of 10 is given when the player has won and a value of -10 when the player loses. A value of 0 is given otherwise, for example when the game is a draw. We note that, in order to improve sample efficiency, in the event of a loss we scale the -10 reward based on the amount of damage the adversary has received. Hence, the final reward on a loss will be: 
\(
-10 \cdot \frac{\text{health}}{\text{max\_health}}
\).
This enhancement will allow the agent to favor moves that damage the opponent, and has resulted in speeding up convergence during training.

\subsubsection{Hyper-parameters and configuration}

The training configuration for our experiments was designed to optimize the performance of the DQN agent across various scenarios. For the DQN hyper-parameters, we set the following (See Table \ref{table:training_hyperparameters}):

\begin{table}[h]
\small
\centering
\caption{DQN Training Hyper-parameters and Configuration}
\label{tab:dqn_hyperparameters}
\begin{tabular}{@{}lc@{}}
\toprule
\textbf{Parameter} & \textbf{Value} \\
\midrule
Replay Buffer Capacity & 3000 \\
Training Batch Size & 64 \\
Training Iterations & 1000 Steps \\
Training Steps per Iteration & 2 \\
Optimizer & ADAM \\
Exploration Strategy & Epsilon-Greedy \\
Epsilon Start & 1.0 \\
Epsilon Final & 0.01 \\
Epsilon Decay Frames & $10^3$ \\
Discount Factor (Gamma) & 0.99 \\
Optimizer Learning Rate & 0.001 \\
Target Network Update Frequency & 1 \\
Temporal Horizon (T\_HORIZON) & 1024 \\
\midrule
\multicolumn{2}{c}{\textbf{Trajectory Generation Methods}} \\
\midrule
Method 1 & Rules-based AI \\
Method 2 & 80\% Rules-based AI + 20\%LLM\\
\bottomrule
\end{tabular}
\label{table:training_hyperparameters}
\end{table}

For for trajectory generation, we use different methods by the adversary/opponent AI, which we aim to compare: (1) Using Rules-based AI, and (2) Using LLMs + Rules-based AI. Note that in the second method, to mitigate the relative slowness (and high cost) of LLMs compared to the Rules-based AI we use a method where only a certain portion of the trajectories generated are LLM based. For method number (2) we employ an 80/20 split where set 20 percent of the trajectories generated are against an LLM opponent. We also evaluate based on the type LLMs used (e.g. Mistral vs LLaMa3 vs GPT-4o).

\subsection{Evaluation Methods}

After training the DQN agents with various types of adversaries we then conduct a 1v1 tournament comprising of 30 rounds each against all types of opponents, which includes all the LLMs used as well as a random action agent to serve as a control. We also look at various combat logs and obtain qualitative insights on the performance of each agent as well as any unusual strategies or actions found.

\section{Results}
We ran the experiments on an Nvidia RTX 3090 and two Nvidia A100 40GB, with each training run taking 4 hours to complete for the Rules-based method to almost a day for those with LLMs based adversaries. Each tournament (fighter only class, all classes) run takes 5 hours each to run. Below are the results of training the RL agent after 1000 epochs (Fig \ref{fig:RL-Methods}):

\begin{figure}[ht]
\centering
\footnotesize
\begin{tikzpicture}
\begin{groupplot}[
    group style={
        group size=2 by 2, 
        vertical sep=1.2cm,
        horizontal sep=1.5cm
    },
    width=0.5\textwidth,
    height=4.0cm,
    ylabel={Average Reward},
    xmin=10, xmax=1000, 
    xtick={10,200,400,600,800,1000}, 
    ymin=-10, ymax=10,
    ytick={-10,-5,0,5,10},
    grid style=dashed,
    legend pos=south east,
    legend style={font=\footnotesize}
]
\nextgroupplot[title={Training vs Rules-based AI}]
\addplot[
    color=red,
    dotted,
    very thick,
]
coordinates {
    (0,5.96) (1000,5.96) 
};
\addlegendentry{best=5.96}
\addplot[
    color=blue,
    thick,
    mark=circle,
    ]
    coordinates {
    (10,-7.634523809523809)(20,-3.7698412698412698)(30,-2.4611111111111112)(40,0.12420634920634925)(50,-4.1460317460317455)(60,-3.848412698412699)(70,-1.9047619047619049)(80,1.5638888888888889)(90,-1.4793650793650794)(100,0.19523809523809518)(110,-1.801984126984127)(120,2.0448412698412697)(130,0.030158730158729982)(140,0.8888888888888887)(150,1.3702380952380953)(160,3.7460317460317456)(170,1.925)(180,-1.6642857142857141)(190,2.0055555555555555)(200,1.753968253968254)(210,1.8)(220,1.5722222222222222)(230,2.4007936507936507)(240,4.449999999999999)(250,3.3579365079365084)(260,1.650793650793651)(270,1.9138888888888885)(280,4.8388888888888895)(290,1.8265873015873018)(300,0.757936507936508)(310,2.0436507936507935)(320,-0.9964285714285714)(330,1.2666666666666668)(340,3.8055555555555554)(350,3.211111111111111)(360,0.1599206349206352)(370,0.38888888888888873)(380,2.65)(390,2.2079365079365085)(400,0.7555555555555554)(410,5.116666666666667)(420,1.7682539682539684)(430,2.522222222222222)(440,3.7888888888888888)(450,4.283333333333333)(460,3.4333333333333336)(470,3.2999999999999994)(480,2.354761904761905)(490,0.5055555555555555)(500,0.041666666666666644)(510,3.0980158730158736)(520,4.307936507936508)(530,3.40952380952381)(540,2.3083333333333336)(550,1.9777777777777776)(560,2.0888888888888886)(570,2.8809523809523814)(580,0.9626984126984126)(590,2.7999999999999994)(600,5.114285714285714)(610,0.24246031746031732)(620,3.538492063492064)(630,3.6317460317460313)(640,3.8337301587301584)(650,3.584920634920635)(660,2.906746031746031)(670,4.142857142857142)(680,0.9460317460317461)(690,4.44404761904762)(700,3.226190476190476)(710,3.2)(720,3.3317460317460315)(730,3.271428571428571)(740,1.1944444444444442)(750,3.425396825396825)(760,2.7376984126984127)(770,5.161904761904761)(780,3.8416666666666663)(790,3.501587301587301)(800,1.734920634920635)(810,2.083333333333333)(820,3.7476190476190476)(830,3.2904761904761903)(840,5.750396825396825)(850,4.102777777777778)(860,2.8706349206349207)(870,2.6519841269841264)(880,3.9178571428571427)(890,1.575)(900,3.088095238095238)(910,4.911111111111111)(920,3.738888888888889)(930,3.4444444444444446)(940,5.966666666666666)(950,3.6869047619047617)(960,5.313888888888889)(970,5.48452380952381)(980,4.851984126984127)(990,2.298809523809524)(1000,4.462301587301587)

    };
    \addlegendentry{Rules-based AI}

\nextgroupplot[title={Training vs Llama 3.1 8B}]
\addplot[
    color=red,
    dotted,
    very thick,
]
coordinates {
    (0,7.27) (1000,7.27) 
};

\addlegendentry{best=7.27}
\addplot[
    color=blue,
    mark=circle,
    ]
    coordinates {
    (10,-7.634523809523809)(20,-3.7698412698412698)(30,-2.4611111111111112)(40,0.12420634920634925)(50,-4.1460317460317455)(60,-3.848412698412699)(70,-1.9047619047619049)(80,1.5638888888888889)(90,-1.4793650793650794)(100,0.19523809523809518)(110,-1.801984126984127)(120,2.0448412698412697)(130,0.030158730158729982)(140,0.8888888888888887)(150,1.3702380952380953)(160,3.7460317460317456)(170,1.925)(180,-1.6642857142857141)(190,2.0055555555555555)(200,1.753968253968254)(210,1.8)(220,1.5722222222222222)(230,2.4007936507936507)(240,4.449999999999999)(250,3.3579365079365084)(260,1.650793650793651)(270,1.9138888888888885)(280,4.8388888888888895)(290,1.8265873015873018)(300,0.757936507936508)(310,2.0436507936507935)(320,-0.9964285714285714)(330,1.2666666666666668)(340,3.8055555555555554)(350,3.211111111111111)(360,0.1599206349206352)(370,0.38888888888888873)(380,2.65)(390,2.2079365079365085)(400,0.7555555555555554)(410,5.116666666666667)(420,1.7682539682539684)(430,2.522222222222222)(440,3.7888888888888888)(450,4.283333333333333)(460,3.4333333333333336)(470,3.2999999999999994)(480,2.354761904761905)(490,0.5055555555555555)(500,0.041666666666666644)(510,3.0980158730158736)(520,4.307936507936508)(530,3.40952380952381)(540,2.3083333333333336)(550,1.9777777777777776)(560,2.0888888888888886)(570,2.8809523809523814)(580,0.9626984126984126)(590,2.7999999999999994)(600,5.114285714285714)(610,0.24246031746031732)(620,3.538492063492064)(630,3.6317460317460313)(640,3.8337301587301584)(650,3.584920634920635)(660,2.906746031746031)(670,4.142857142857142)(680,0.9460317460317461)(690,4.44404761904762)(700,3.226190476190476)(710,3.2)(720,3.3317460317460315)(730,3.271428571428571)(740,1.1944444444444442)(750,3.425396825396825)(760,2.7376984126984127)(770,5.161904761904761)(780,3.8416666666666663)(790,3.501587301587301)(800,1.734920634920635)(810,2.083333333333333)(820,3.7476190476190476)(830,3.2904761904761903)(840,5.750396825396825)(850,4.102777777777778)(860,2.8706349206349207)(870,2.6519841269841264)(880,3.9178571428571427)(890,1.575)(900,3.088095238095238)(910,4.911111111111111)(920,3.738888888888889)(930,3.4444444444444446)(940,5.966666666666666)(950,3.6869047619047617)(960,5.313888888888889)(970,5.48452380952381)(980,4.851984126984127)(990,2.298809523809524)(1000,4.462301587301587)

    };
\addplot[
    color=black,
    thick,
    mark=circle,
    ]
    coordinates {
        (10,-3.9793650793650794)(20,-3.2)(30,0.006349206349206469)(40,-0.5111111111111112)(50,1.4134920634920636)(60,1.16984126984127)(70,4.276984126984127)(80,3.897222222222222)(90,3.683333333333333)(100,4.602777777777778)(110,4.719444444444445)(120,5.722619047619048)(130,3.481746031746032)(140,1.1047619047619048)(150,2.291666666666667)(160,0.4865079365079365)(170,3.116269841269841)(180,3.1488095238095237)(190,2.1234126984126984)(200,2.209920634920635)(210,1.1349206349206349)(220,5.747222222222222)(230,4.988888888888889)(240,0.024206349206349172)(250,-2.3694444444444445)(260,1.2611111111111108)(270,5.344444444444445)(280,4.127777777777777)(290,3.08968253968254)(300,4.7555555555555555)(310,2.6238095238095243)(320,2.036111111111111)(330,5.841666666666668)(340,4.027777777777778)(350,7.272222222222222)(360,2.915079365079366)(370,3.5023809523809524)(380,6.705555555555554)(390,3.443650793650794)(400,2.738888888888889)(410,3.0218253968253963)(420,3.6277777777777778)(430,6.666666666666668)(440,1.9988095238095238)(450,2.4555555555555553)(460,5.213492063492064)(470,1.8948412698412698)(480,0.9869047619047621)(490,2.572619047619048)(500,3.999206349206349)(510,4.097222222222222)(520,2.5567460317460315)(530,3.0853174603174605)(540,4.417857142857144)(550,-0.8222222222222222)(560,4.993650793650794)(570,3.4972222222222222)(580,5.073015873015874)(590,5.2174603174603185)(600,3.5253968253968253)(610,5.95)(620,1.791666666666667)(630,1.4194444444444443)(640,3.851190476190476)(650,4.477777777777778)(660,6.091666666666666)(670,5.3547619047619035)(680,4.166666666666667)(690,3.6003968253968255)(700,3.8944444444444444)(710,4.131746031746032)(720,2.3583333333333334)(730,5.3)(740,4.355555555555555)(750,5.263888888888889)(760,3.166666666666667)(770,4.233333333333333)(780,3.393253968253968)(790,5.10952380952381)(800,5.288888888888889)(810,5.366666666666667)(820,1.8444444444444446)(830,6.559523809523809)(840,5.111111111111111)(850,3.477777777777778)(860,6.316666666666667)(870,2.963492063492063)(880,5.363888888888889)(890,3.061111111111111)(900,5.522619047619048)(910,5.085714285714285)(920,4.78968253968254)(930,3.151587301587301)(940,3.7107142857142854)(950,4.594444444444445)(960,4.086507936507937)(970,4.294444444444444)(980,4.1)(990,3.95)(1000,3.661111111111111)

    };
    \addlegendentry{Llama 3.1 8B}

\nextgroupplot[title={Training vs Mistral-0.3}]
\addplot[
    color=red,
    dotted,
    very thick,
]
coordinates {
    (0,7.92) (1000,7.92) 
};

\addlegendentry{best=7.92}
\addplot[
    color=blue,
    mark=circle,
    ]
    coordinates {
(1,-7.634523809523809)(2,-3.7698412698412698)(3,-2.4611111111111112)    (10,-7.634523809523809)(20,-3.7698412698412698)(30,-2.4611111111111112)(40,0.12420634920634925)(50,-4.1460317460317455)(60,-3.848412698412699)(70,-1.9047619047619049)(80,1.5638888888888889)(90,-1.4793650793650794)(100,0.19523809523809518)(110,-1.801984126984127)(120,2.0448412698412697)(130,0.030158730158729982)(140,0.8888888888888887)(150,1.3702380952380953)(160,3.7460317460317456)(170,1.925)(180,-1.6642857142857141)(190,2.0055555555555555)(200,1.753968253968254)(210,1.8)(220,1.5722222222222222)(230,2.4007936507936507)(240,4.449999999999999)(250,3.3579365079365084)(260,1.650793650793651)(270,1.9138888888888885)(280,4.8388888888888895)(290,1.8265873015873018)(300,0.757936507936508)(310,2.0436507936507935)(320,-0.9964285714285714)(330,1.2666666666666668)(340,3.8055555555555554)(350,3.211111111111111)(360,0.1599206349206352)(370,0.38888888888888873)(380,2.65)(390,2.2079365079365085)(400,0.7555555555555554)(410,5.116666666666667)(420,1.7682539682539684)(430,2.522222222222222)(440,3.7888888888888888)(450,4.283333333333333)(460,3.4333333333333336)(470,3.2999999999999994)(480,2.354761904761905)(490,0.5055555555555555)(500,0.041666666666666644)(510,3.0980158730158736)(520,4.307936507936508)(530,3.40952380952381)(540,2.3083333333333336)(550,1.9777777777777776)(560,2.0888888888888886)(570,2.8809523809523814)(580,0.9626984126984126)(590,2.7999999999999994)(600,5.114285714285714)(610,0.24246031746031732)(620,3.538492063492064)(630,3.6317460317460313)(640,3.8337301587301584)(650,3.584920634920635)(660,2.906746031746031)(670,4.142857142857142)(680,0.9460317460317461)(690,4.44404761904762)(700,3.226190476190476)(710,3.2)(720,3.3317460317460315)(730,3.271428571428571)(740,1.1944444444444442)(750,3.425396825396825)(760,2.7376984126984127)(770,5.161904761904761)(780,3.8416666666666663)(790,3.501587301587301)(800,1.734920634920635)(810,2.083333333333333)(820,3.7476190476190476)(830,3.2904761904761903)(840,5.750396825396825)(850,4.102777777777778)(860,2.8706349206349207)(870,2.6519841269841264)(880,3.9178571428571427)(890,1.575)(900,3.088095238095238)(910,4.911111111111111)(920,3.738888888888889)(930,3.4444444444444446)(940,5.966666666666666)(950,3.6869047619047617)(960,5.313888888888889)(970,5.48452380952381)(980,4.851984126984127)(990,2.298809523809524)(1000,4.462301587301587)
    };

\addplot[
    color=black,
    thick,
    mark=circle,
    ]
    coordinates {
   (10,-6.290873015873015)(20,-2.0277777777777777)(30,0.11706349206349195)(40,2.688888888888889)(50,3.613888888888889)(60,2.321825396825397)(70,-2.0801587301587303)(80,6.738888888888889)(90,5.861111111111111)(100,-0.28888888888888886)(110,0.4722222222222222)(120,2.897222222222222)(130,1.8702380952380953)(140,3.215079365079365)(150,4.697222222222222)(160,4.128968253968254)(170,3.208333333333333)(180,0.37539682539682545)(190,5.330555555555556)(200,4.794444444444444)(210,5.791666666666666)(220,2.3083333333333336)(230,3.3186507936507934)(240,4.4714285714285715)(250,2.8499999999999996)(260,3.075396825396825)(270,0.3563492063492063)(280,5.708730158730159)(290,2.563888888888889)(300,6.094047619047619)(310,2.555952380952381)(320,0.5555555555555555)(330,6.249603174603175)(340,3.080952380952381)(350,2.28452380952381)(360,4.169444444444444)(370,0.488888888888889)(380,5.575)(390,3.4972222222222222)(400,1.1527777777777777)(410,4.058333333333334)(420,5.048412698412699)(430,1.0333333333333334)(440,2.5202380952380947)(450,5.280555555555555)(460,3.983333333333334)(470,1.8642857142857145)(480,2.9424603174603177)(490,4.813888888888889)(500,3.5305555555555554)(510,1.775)(520,5.5)(530,2.9309523809523808)(540,3.7869047619047613)(550,4.986111111111112)(560,5.692857142857143)(570,5.222222222222221)(580,6.666666666666666)(590,0.520238095238095)(600,4.1)(610,5.566666666666666)(620,2.413888888888889)(630,2.4845238095238096)(640,3.413888888888889)(650,3.8940476190476194)(660,4.763888888888889)(670,3.7063492063492065)(680,6.102777777777778)(690,4.677777777777778)(700,1.6238095238095238)(710,2.608333333333333)(720,2.073809523809524)(730,4.083333333333333)(740,2.958333333333334)(750,3.303571428571429)(760,5.258333333333334)(770,3.598412698412699)(780,6.4527777777777775)(790,4.8388888888888895)(800,2.0253968253968258)(810,5.169444444444446)(820,5.012698412698413)(830,3.0305555555555554)(840,4.8805555555555555)(850,6.6)(860,4.791666666666666)(870,5.111904761904762)(880,4.534523809523809)(890,5.47420634920635)(900,7.927777777777778)(910,5.319444444444445)(920,2.303174603174603)(930,2.575)(940,2.197619047619048)(950,7.933333333333334)(960,2.1055555555555556)(970,4.708333333333334)(980,2.554761904761905)(990,5.941666666666667)(1000,4.908333333333334)

    };
     \addlegendentry{Mistral-0.3}
    
\nextgroupplot[title={Training vs gpt-4o-mini}]

\addplot[
    color=red,
    dotted,
    very thick,
]
coordinates {
    (0,8.00694444444444) (1000,8.00694444444444) 
};

\addlegendentry{best=8.01}

\addplot[
    color=blue,
    mark=circle,
    ]
    coordinates {
    (10,-7.634523809523809)(20,-3.7698412698412698)(30,-2.4611111111111112)(40,0.12420634920634925)(50,-4.1460317460317455)(60,-3.848412698412699)(70,-1.9047619047619049)(80,1.5638888888888889)(90,-1.4793650793650794)(100,0.19523809523809518)(110,-1.801984126984127)(120,2.0448412698412697)(130,0.030158730158729982)(140,0.8888888888888887)(150,1.3702380952380953)(160,3.7460317460317456)(170,1.925)(180,-1.6642857142857141)(190,2.0055555555555555)(200,1.753968253968254)(210,1.8)(220,1.5722222222222222)(230,2.4007936507936507)(240,4.449999999999999)(250,3.3579365079365084)(260,1.650793650793651)(270,1.9138888888888885)(280,4.8388888888888895)(290,1.8265873015873018)(300,0.757936507936508)(310,2.0436507936507935)(320,-0.9964285714285714)(330,1.2666666666666668)(340,3.8055555555555554)(350,3.211111111111111)(360,0.1599206349206352)(370,0.38888888888888873)(380,2.65)(390,2.2079365079365085)(400,0.7555555555555554)(410,5.116666666666667)(420,1.7682539682539684)(430,2.522222222222222)(440,3.7888888888888888)(450,4.283333333333333)(460,3.4333333333333336)(470,3.2999999999999994)(480,2.354761904761905)(490,0.5055555555555555)(500,0.041666666666666644)(510,3.0980158730158736)(520,4.307936507936508)(530,3.40952380952381)(540,2.3083333333333336)(550,1.9777777777777776)(560,2.0888888888888886)(570,2.8809523809523814)(580,0.9626984126984126)(590,2.7999999999999994)(600,5.114285714285714)(610,0.24246031746031732)(620,3.538492063492064)(630,3.6317460317460313)(640,3.8337301587301584)(650,3.584920634920635)(660,2.906746031746031)(670,4.142857142857142)(680,0.9460317460317461)(690,4.44404761904762)(700,3.226190476190476)(710,3.2)(720,3.3317460317460315)(730,3.271428571428571)(740,1.1944444444444442)(750,3.425396825396825)(760,2.7376984126984127)(770,5.161904761904761)(780,3.8416666666666663)(790,3.501587301587301)(800,1.734920634920635)(810,2.083333333333333)(820,3.7476190476190476)(830,3.2904761904761903)(840,5.750396825396825)(850,4.102777777777778)(860,2.8706349206349207)(870,2.6519841269841264)(880,3.9178571428571427)(890,1.575)(900,3.088095238095238)(910,4.911111111111111)(920,3.738888888888889)(930,3.4444444444444446)(940,5.966666666666666)(950,3.6869047619047617)(960,5.313888888888889)(970,5.48452380952381)(980,4.851984126984127)(990,2.298809523809524)(1000,4.462301587301587)
    };
    
    \addplot[
       color=black,
       thick,
       mark=circle,
    ]
    coordinates {
    (10,-3.868055555555556)
    (20,-1.878968253968254)
    (30,-4.171626984126984)
    (40,-2.55489417989418)
    (50,-4.1590608465608465)
    (60,-3.275132275132275)
    (70,0.9206349206349206)
    (80,0.9596560846560845)
    (90,1.0476190476190472)
    (100,4.031746031746032)
    (110,3.5572089947089944)
    (120,3.8118386243386246)
    (130,3.5644841269841274)
    (140,2.1875)
    (150,3.127314814814815)
    (160,2.256944444444444)
    (170,2.7837301587301586)
    (180,1.4004629629629628)
    (190,4.571759259259259)
    (200,2.025462962962963)
    (210,4.493055555555556)
    (220,4.8128306878306875)
    (230,4.388888888888889)
    (240,0.6031746031746031)
    (250,5.689814814814814)
    (260,4.180555555555555)
    (270,3.55489417989418)
    (280,1.9163359788359788)
    (290,3.9652777777777777)
    (300,5.497354497354497)
    (310,3.493055555555556)
    (320,3.9285714285714284)
    (330,2.3888888888888884)
    (340,5.41170634920635)
    (350,4.890542328042328)
    (360,4.104166666666667)
    (370,3.9794973544973544)
    (380,2.7986111111111116)
    (390,2.6134259259259256)
    (400,1.229828042328042)
    (410,2.2916666666666665)
    (420,2.0767195767195767)
    (430,2.275132275132275)
    (440,3.998015873015873)
    (450,5.043981481481481)
    (460,5.212962962962963)
    (470,4.541666666666667)
    (480,4.4692460317460325)
    (490,4.00462962962963)
    (500,3.5671296296296293)
    (510,5.476851851851851)
    (520,1.3865740740740737)
    (530,3.576388888888889)
    (540,4.208333333333333)
    (550,3.8052248677248675)
    (560,3.812830687830688)
    (570,5.321759259259258)
    (580,3.8125)
    (590,5.431547619047619)
    (600,6.986111111111111)
    (610,3.235119047619048)
    (620,3.826058201058201)
    (630,2.2400793650793647)
    (640,4.10978835978836)
    (650,4.726851851851851)
    (660,4.513888888888889)
    (670,4.479166666666667)
    (680,4.55489417989418)
    (690,2.7384259259259256)
    (700,3.4077380952380953)
    (710,2.4533730158730154)
    (720,4.793981481481482)
    (730,3.3280423280423284)
    (740,8.006944444444445)
    (750,3.8435846560846563)
    (760,3.6875000000000004)
    (770,5.2976190476190474)
    (780,7.243055555555556)
    (790,6.261574074074074)
    (800,5.231481481481482)
    (810,5.273148148148148)
    (820,4.32175925925926)
    (830,6.069444444444445)
    (840,5.150462962962964)
    (850,1.636574074074074)
    (860,4.462962962962963)
    (870,5.090277777777777)
    (880,3.3224206349206344)
    (890,6.406415343915344)
    (900,6.738756613756613)
    (910,5.552910052910053)
    (920,2.693783068783069)
    (930,2.054563492063492)
    (940,5.2440476190476195)
    (950,3.357473544973545)
    (960,1.2718253968253967)
    (970,5.279761904761904)
    (980,3.2490079365079367)
    (990,4.583333333333333)
    (1000,4.583333333333333)

    };
    
\addlegendentry{gpt-4o-mini}

\end{groupplot}
\end{tikzpicture}
\caption{Comparison of average rewards of training an RL agent against different LLMs as adversary}
\label{fig:RL-Methods}
\end{figure}
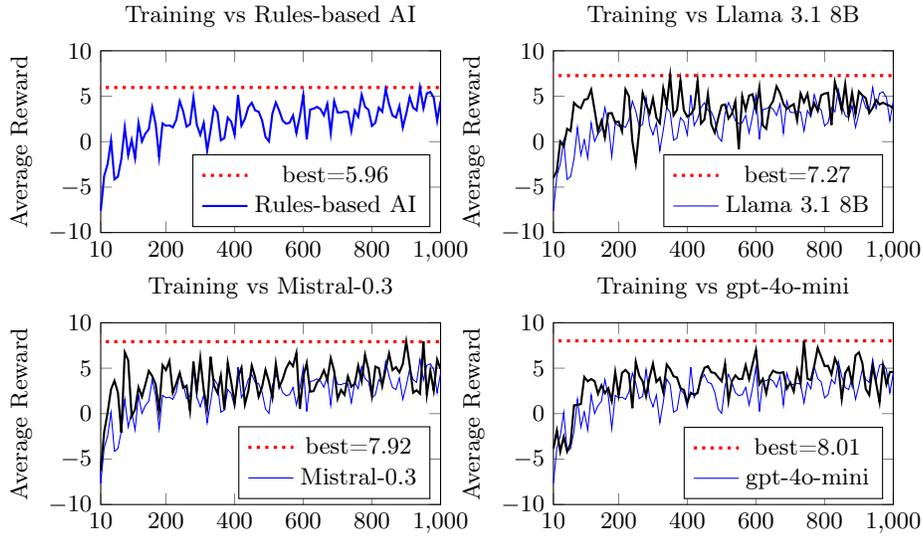

Based on the graphs (Fig \ref{fig:RL-Methods}), training runs using LLMs appear to converge more quickly, also we can see that the best rewards are being obtained by those training against LLM adversaries. 

\subsection{Tournament Results}
We move on to the different LLM vs DQN experiments. Using the trained weights for the RL agents we now pit them in a tournament involving all the LLMs and the rules based AI. We run 30 matches for each player/adversary combination. Two tournaments will be conducted, a fighter-class only tournament and one with all four classes involved. Agents that start with \textbf{llm\_} are agents that depend entirely on the LLM for their policy while those with \textbf{rl\_} are agents trained using DQN based on a neural network (Fig \ref{fig:dqn-network}) with their adversary controlled by the indicated LLM or the rules-based AI. Below is the result of each agent having completed 240 games across all the other agents.

\begin{table}[h]
\centering
\caption{D\&D 5E Fighter-Class Tournament: Win/Loss/Tie Matrix}
\resizebox{\textwidth}{!}{%
\begin{tabular}{|l|c|c|c|c|c|c|c|c|c|}
\hline
\textbf{Agent} & \textbf{AI rules-based} & \textbf{llm\_gpt4} & \textbf{llm\_llama3} & \textbf{llm\_mistral} & \textbf{random} & \textbf{rl\_gpt4\_trained} & \textbf{rl\_llama3\_trained} & \textbf{rl\_mistral\_trained} & \textbf{rl\_rules\_trained} \\
\hline
AI rules-based& -  & 11/19/0 & 16/14/0 & 21/9/0 & 29/1/0 & 10/17/3 & 21/9/0 & 12/18/0 & 7/22/1 \\
llm\_gpt4 & 19/11/0 & -  & 20/10/0 & 5/25/0 & 28/2/0 & 14/16/0 & 19/11/0 & 21/9/0 & 16/11/3 \\
llm\_llama3 & 14/16/0 & 10/20/0 & -  & 5/25/0 & 26/3/1 & 13/17/0 & 17/13/0 & 16/14/0 & 12/17/1 \\
llm\_mistral & 9/21/0 & 25/5/0 & 25/5/0 & -  & 29/1/0 & 20/10/0 & 22/8/0 & 19/9/2 & 25/4/1 \\
random & 1/29/0 & 2/28/0 & 3/26/1 & 1/29/0 & -  & 1/29/0 & 1/29/0 & 3/22/5 & 2/23/5 \\
rl\_gpt4\_trained & 17/10/3 & 16/14/0 & 17/13/0 & 10/20/0 & 29/1/0 & -  & 18/7/5 & 12/13/5 & 13/6/11 \\
rl\_llama3\_trained & 9/21/0 & 11/19/0 & 13/17/0 & 8/22/0 & 29/1/0 & 7/18/5 & -  & 10/11/9 & 13/9/8 \\
rl\_mistral\_trained & 18/12/0 & 9/21/0 & 14/16/0 & 9/19/2 & 22/3/5 & 13/12/5 & 11/10/9 & -  & 13/5/12 \\
rl\_rules\_trained & 22/7/1 & 11/16/3 & 17/12/1 & 4/25/1 & 23/2/5 & 6/13/11 & 9/13/8 & 5/13/12 & -  \\

\hline
\end{tabular}%
}

\label{tab:fighter-class-matrix}
\end{table}

\begin{table}[h]
\centering

{\footnotesize 
\caption{Leaderboard for the Fighter-Class Tournament: LLMs vs RL Agents}
\begin{tabular}{|l|c|c|c|c|}
\hline
\textbf{Agent} & \textbf{Wins} & \textbf{Losses} & \textbf{Ties} & \textbf{AVG Rounds} \\
\hline

llm\_mistral & 174.0 & 63.0 & 3.0 & 56.63 \\
llm\_gpt4 & 142.0 & 95.0 & 3.0 & 52.77 \\
rl\_gpt4\_trained & 132.0 & 84.0 & 24.0 & 188.47 \\
AI (rules-based) & 127.0 & 109.0 & 4.0 & 50.60 \\
llm\_llama3 & 113.0 & 125.0 & 2.0 & 60.40 \\
rl\_mistral\_trained & 109.0 & 98.0 & 33.0 & 176.10 \\
rl\_llama3\_trained & 100.0 & 118.0 & 22.0 & 161.57 \\
rl\_rules\_trained & 97.0 & 101.0 & 42.0 & 264.07 \\
random & 14.0 & 215.0 & 11.0 & 100.33 \\

\hline
\end{tabular}
} 

\label{tab:fighter-class-leaderboard}
\end{table}

LLM models, particularly \textbf{llm\_mistral} and \textbf{llm\_gpt4}, performed very well, taking the top and 2nd spots respectively. The RL models showed mixed results, with \textbf{rl\_gpt4\_trained} performing exceptionally well (3rd place), while others were more middle-of-the-pack. The 'AI' rules based model, landing in 4th place while the random agent, as expected, performed poorly, confirming that the other agents are indeed employing some form of strategy.

Next, the same process is repeated but we now allow an additional three classes namely Rogue, Mage and cleric which significantly makes the environment more complicated.

\begin{table}[h!]
\centering
\caption{D\&D 5E Four Classes Tournament: Win/Loss/Tie Matrix}
\resizebox{\textwidth}{!}{%
\begin{tabular}{|l|c|c|c|c|c|c|c|c|c|}
\hline
\textbf{Agent} & \textbf{ai} & \textbf{llm\_gpt4} & \textbf{llm\_llama3} & \textbf{llm\_mistral} & \textbf{random} & \textbf{rl\_gpt4\_trained} & \textbf{rl\_llama3\_trained} & \textbf{rl\_mistral\_trained} & \textbf{rl\_rules\_trained} \\
\hline
ai & -  & 16/14/0  & 19/11/0  & 17/13/0  & 23/7/0  & 17/12/1  & 13/17/0  & 6/24/0  & 12/17/1  \\
llm\_gpt4 & 14/16/0  & -  & 20/10/0  & 18/12/0  & 19/11/0  & 13/17/0  & 9/20/1  & 15/14/1  & 8/21/1  \\
llm\_llama3 & 11/19/0  & 10/20/0  & -  & 11/19/0  & 14/12/4  & 8/21/1  & 9/19/2  & 7/18/5  & 9/21/0  \\
llm\_mistral & 13/17/0  & 12/18/0  & 19/11/0  & -  & 23/7/0  & 5/24/1  & 8/21/1  & 14/13/3  & 13/17/0  \\
random & 7/23/0  & 11/19/0  & 12/14/4  & 7/23/0  & -  & 6/22/2  & 5/23/2  & 7/23/0  & 6/22/2  \\
rl\_gpt4\_trained & 12/17/1  & 17/13/0  & 21/8/1  & 24/5/1  & 22/6/2  & -  & 21/7/2  & 17/13/0  & 12/17/1  \\
rl\_llama3\_trained & 17/13/0  & 20/9/1  & 19/9/2  & 21/8/1  & 23/5/2  & 7/21/2  & -  & 9/19/2  & 15/10/5  \\
rl\_mistral\_trained & 24/6/0  & 14/15/1  & 18/7/5  & 13/14/3  & 23/7/0  & 13/17/0  & 19/9/2  & -  & 18/10/2  \\
rl\_rules\_trained & 17/12/1  & 21/8/1  & 21/9/0  & 17/13/0  & 22/6/2  & 17/12/1  & 10/15/5  & 10/18/2  & -  \\

\hline
\end{tabular}%
}

\label{tab:dnd-four-classes-matrix}
\end{table}

\begin{table}[h]
\centering
\caption{Leaderboard for D\&D 5E Four Classes Tournament: LLMs vs RL Agents}
{\footnotesize 
\begin{tabular}{|l|c|c|c|c|}
\hline
\textbf{Agent} & \textbf{Wins} & \textbf{Losses} & \textbf{Ties} & \textbf{AVG Rounds} \\
\hline

rl\_gpt4\_trained & 146.0 & 86.0 & 8.0 & 49.30 \\
rl\_mistral\_trained & 142.0 & 85.0 & 13.0 & 66.50 \\
rl\_rules\_trained & 135.0 & 93.0 & 12.0 & 77.57 \\
rl\_llama3\_trained & 131.0 & 94.0 & 15.0 & 86.93 \\
ai (rules-based) & 123.0 & 115.0 & 2.0 & 28.60 \\
llm\_gpt4 & 116.0 & 121.0 & 3.0 & 42.87 \\
llm\_mistral & 107.0 & 128.0 & 5.0 & 37.90 \\
llm\_llama3 & 79.0 & 149.0 & 12.0 & 52.07 \\
random & 61.0 & 169.0 & 10.0 & 53.80 \\

\hline
\end{tabular}
}

\label{tab:dnd-four-classes-leaderboard}
\end{table}

The RL models, particularly \textbf{rl\_pt4\_trained}, \textbf{rl\_llama3\_trained} and \textbf{rl\_mistral\_trained}, have risen to the top of the leaderboard. LLM models, while still performing well, are no longer dominating the top spots as they did in the fighter-only tournament. The rules-based AI has moved just above of the LLMs in standing, this highlights the challenges of building Rules-based AI as the environment becomes increasingly complex, though a properly written one can still be better than LLM only models.

The RL models show improved performance across different matchups, suggesting they adapt well to the increased complexity of multiple character classes.
\textbf{rl\_llama3\_trained} and \textbf{rl\_mistral\_trained} consistently perform well against various opponents, including the LLMs by themselves and the rule-based AI.

The average round length has also decreased, which probably indicates various imbalances in the matches due to the other classes having lower hit points as compared to a fighter class.

It is important to note that, except for the LLaMA3-trained one, the RL agents trained using LLM adversaries outperform the rules based adversaries indicating that using LLMs improves the training process, however it's necessary to evaluate prompting strategies in order to improve the stability of responses. The GPT-4o based LLM with its function to use tools has shown to provide the most stable response and hence has improved performance versus the other LLMs. LLaMA3 in contrast does not respond well to the same prompt and results in providing invalid responses, for example while it was able to explain its decision logically, it could not provide the correct index for the action. A different prompting and response parsing strategy should be further explored to improve on this as well as fine tuning the LLM to obtain a response stable variant.

\section{Discussion}

The results of the D\&D 5E four classes tournament reveal significant changes in the performance hierarchy compared to the fighter-only tournament:
\begin{itemize}
    \item \textbf{RL Model Dominance:}
    The top three positions are occupied by RL models  \text{rl\_gpt4\_trained}, \text{rl\_mistral\_trained}, \text{rl\_rules\_trained} and \text{rl\_lama3\_trained}. This suggests that RL approaches are particularly effective at handling the increased complexity introduced by multiple character classes.

    \item \textbf{LLM Performance:}
    LLMs showed mixed results. While LLMs like  \text{llm\_gpt4} and \text{llm\_mistral} performed well in the fighter-only tournament. This changed in the 4-Class tournament where they were at the bottom of the  leaderboard. This indicates that while LLMs can be effective, they may struggle more with the increased complexity and specific rule interactions of multiple classes.

    \item \textbf{Consistency of RL Approaches:}
    The consistent high performance of LLM-adversary RL models across different architectures (Mistral, GPT-4o, LLaMA3) indicates that the RL training process is robust and effective for this task, regardless of the underlying LLM used. However, this does not mean that the quality of the LLM should be ignored, as some agents perform better depending on which LLM they trained against. This can be seen with the fact that the RL agents mirror the positions of their corresponding LLM teachers in the leaderboard.

    \item \textbf{Superior performance of LLM adversary trained RL agents:}
    In both tournaments, we see that LLM trained RL agents outperformed the pure rule based adversary trained ones.
\end{itemize}

In order to explain the advantages of LLMs as an adversary we checked the combat logs and noticed that "bugs" in the rules-based AI get consistently taken advantage of during the RL training process, this exploitation of behavior does not provide a strong signal for the agent to perform well in the real world.

Although LLMs exhibit significant capabilities in understanding and generating natural language, their application in real-time strategic decision-making environments proves challenging. While LLMs on their own lack the consistency needed to perform well in strategic decision-making, using it as a means to guide RL agents by providing adversarial feedback makes them more effective in this regard.

The inherent complexity of LLMs in processing language and generating contextually appropriate responses, though beneficial for many natural language processing (NLP) tasks, it poses limitations in fast-paced strategic environments. This slower decision-making process impacts their effectiveness in real-time scenarios. Despite this, LLMs showed promise in their ability to develop and adapt strategies over time, albeit at a slower rate compared to RL agents.

\section{Conclusions and Future Work}

This study demonstrates the potential and challenges of integrating LLMs into RL environments, specifically within the context of D\&D 5E combat scenarios. We successfully created an open-source D\&D 5E combat framework in Python and set up vLLM to mimic the OpenAI API, enabling us to train an RL agent using LLMs. The findings suggest that LLM-trained RL agents typically outperform both LLM-controlled adversaries and rule-based AI agents. However, in more complex environments, LLMs used as adversaries have demonstrated to be more effective teachers than rule-based AI systems.

Moving forward, several avenues should be explored to build on these findings. Optimizing the integration of LLMs within RL frameworks is crucial; Developing methods to reduce LLMs' decision-making time and fine-tuning models within the same series, with careful attention to data selection, success metrics, and scalability, will optimize performance while preserving their strategic depth. Also, enhancing the framework to incorporate multi-agent training in order to capture cooperation in a multiplayer D\&D 5E game where there are multiple party members all cooperating to defeat multiple opponents and solve various tasks. Additionally, investigating hybrid models that combine the strengths of RL agents and LLMs could lead to more robust and adaptable AI systems. Testing these integrated models within more complex and varied scenarios within D\&D 5E and other strategic games will further validate their capabilities. Improving the sample efficiency of RL agents is also important to ensure faster learning and adaptation to new environments. Furthermore, exploring additional RL models, LLM prompting techniques, and incorporating more complex environmental factors such as different classes, actions, items, and battlegrounds will enrich the research. Finally, leveraging the developed framework to create educational tools for AI training could provide dynamic and unpredictable simulation environments for students and researchers, and foster a deeper understanding of AI-driven strategic decision-making.
\begin{credits}

\subsubsection{\discintname}
The authors have no competing interests to disclose.

\end{credits}

\bibliographystyle{splncs04}
\bibliography{1357.bib}
\end{document}